\documentclass[runningheads]{llncs}
\usepackage{todonotes}
\usepackage{amsmath}
\usepackage{graphicx}
\usepackage{comment}
\usepackage{caption}
\usepackage{subcaption}
\usepackage{multirow}
\usepackage{cite}
\usepackage{booktabs}

\usepackage{array}
\usepackage{amssymb}
\usepackage{placeins}
\usepackage{soul}
\usepackage{url}

\begin{document}

\title{The Performance of Transferability Metrics\\does not Translate to Medical Tasks}

\author{Levy Chaves\inst{1} \and Alceu Bissoto\inst{1} \and
Eduardo Valle\inst{2,3} \and Sandra Avila\inst{1}}

\authorrunning{Chaves et al.}

\institute{Recod.ai Lab, Institute of Computing, University of Campinas, Brazil \email{\{levy.chaves, alceubissoto, sandra\}@ic.unicamp.br}
\and School of Electrical and Computing Engineering, University of Campinas, Brazil \email{dovalle@dca.fee.unicamp.br} \and Valeo.ai Paris}
%

\maketitle              
\begin{abstract}
Transfer learning boosts the performance of medical image analysis by enabling deep learning (DL) on small datasets through the knowledge acquired from large ones. As the number of DL architectures explodes, exhaustively attempting all candidates becomes unfeasible, motivating cheaper alternatives for choosing them. Transferability scoring methods emerge as an enticing solution, allowing to efficiently calculate a score that correlates with the architecture accuracy on any target dataset. However, since transferability scores have not been evaluated on medical datasets, their use in this context remains uncertain, preventing them from benefiting practitioners. We fill that gap in this work, thoroughly evaluating seven transferability scores in three medical applications, including out-of-distribution scenarios. Despite promising results in general-purpose datasets, our results show that no transferability score can reliably and consistently estimate target performance in medical contexts, inviting further work in~that~direction.

\end{abstract}
\section{Introduction}

\begin{figure}[htb!]
    \centering
    \includegraphics[width=\textwidth]{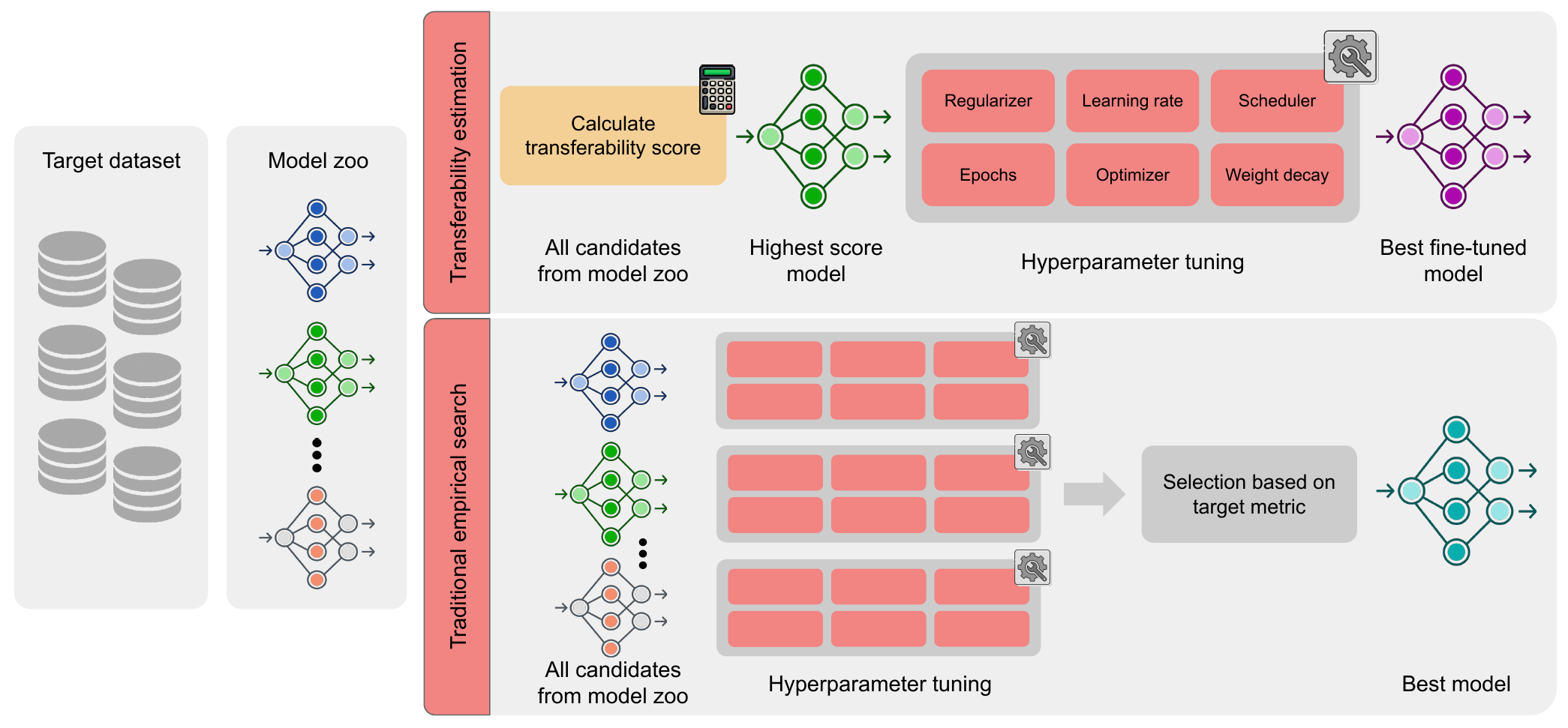}
    \caption{Transferability estimation vs.~traditional empirical search. The latter selects the best candidate model through empirical evaluation of the target metric, thus needing a costly hyperparameter search for each candidate model. Transferability computes instead a proxy score that correlates with the best expected fine-tuned performance. Only the selected model (highest score) will need hyperparameter tuning to obtain the optimized model.}
    \label{fig:transferability_overview}
\end{figure}

Transfer learning allows, in data-limited scenarios, to leverage knowledge obtained from larger datasets. Due to its effectiveness, it is the preferred training method in medical image analysis~\cite{matsoukas2022makes}. Practitioners typically fine-tune a pre-trained model for the target task. However, selecting the most appropriate pre-trained model can significantly impact the final performance. The growing number of architectures and datasets has led to increasingly difficult decisions. While, with unlimited resources, it would be theoretically possible to compare all options empirically, that approach is too inefficient in practice. Sound empirical evaluation must often be tempered with the designer's experience and, often, not-so-sound intuition, prejudices, and hearsay.

Transferability estimation promises to ease this burden, as shown in Fig.~\ref{fig:transferability_overview}. Traditional empirical selection of architectures requires optimizing the hyper-parameters of each candidate to allow a fair comparison~\cite{tuningplaybookgithub}. Transferability scoring methods, in contrast, allow efficiently selecting the most promising model for a~given target dataset without fine-tuning each candidate. When the transferability score accurately measures the ability to transfer knowledge between arbitrary tasks, empirical comparison of models may be limited to a small subset of candidates.  

\begin{figure}[htb!]
\centering

\includegraphics[width=\textwidth]{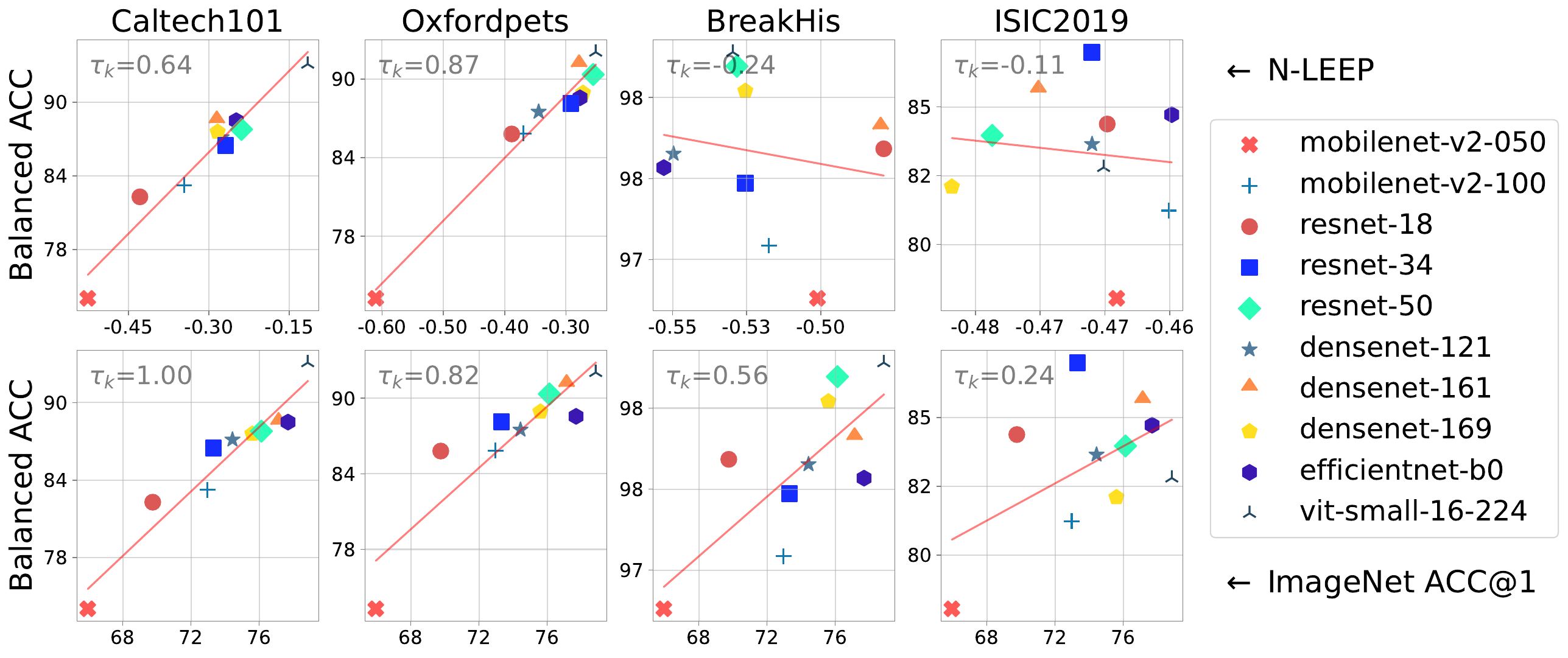}
\caption{Both our best transferability score (N-LEEP) and the ImageNet ACC@1 metric (baseline) on the source model suffice to predict performance on target general-purpose tasks (two left columns). For target medical tasks, the scenario is much different, with neither scores nor raw performances being strong predictors of transferability (two right columns). A good transferability scorer should capture how well a transferability score (x-axis) relates to a test performance metric (y-axis), i.e, higher values of transferability scores predict higher values of true performance. The red line showcases any linear trending between the scores and the accuracy on the task for a given source model. }
\label{fig:sample-natural-vs-medical-scores}
\end{figure}

Transferability scoring methods have shown promising results, performing well when source and target datasets share strong similarities in classes and image characteristics~\cite{ibrahim2021newer,agostinelli2022stable}. However, as we will see, their behavior is much different for target medical datasets, a situation in which the target dataset deviates much more intensely from the source dataset as depicted in Fig.~\ref{fig:sample-natural-vs-medical-scores}.

This work evaluates several transferability scoring methods in the medical domain, including skin lesions, brain tumors, and breast cancer. We define a comprehensive hyperparameter optimization to ensure that the fine-tuned models are evaluated on their best capabilities. Additionally, we extend the evaluation to investigate how transferability scores correlate with out-of-distribution performance. We include at least one dataset considered out-of-distribution from a source one for each medical application.

In summary, the contributions of our paper are twofold:
\begin{itemize}
    \item We extensively evaluate seven transferability scoring methods for three distinct medical classification tasks, covering common imagery types in medical~tasks; 
    \item We design a new methodology for the medical context to account for out-of-distribution evaluation of transferability scoring methods.
\end{itemize}

\section{Transferability Scores \& Related Work}
\label{sec:concepts}

An effective transferability scoring method exhibits computational efficiency while strongly correlating with the final performance metric of a fine-tuned model on the target dataset. Generally, the estimation of transferability involves extracting the embeddings or predictions from the target dataset. That extracted information is integrated with the target dataset's ground-truth labels to quantify the model's transferability. Transferability scoring methods can be categorized into feature-based (fb) and source-label-based (sb). Source-label-based scores assume access to the source classification head for calculating probability distribution or label predictions, whereas feature-based scores only require source models for feature extraction. Both methods require the true labels of the target dataset for computing the transferability score. We summarize the transferability scoring methods, sorted by publication date in Table~\ref{tab:transferability-metrics}.

\begin{table}[t]
\centering
\caption{Summary of transferability scoring methods (Tr. scorer), sorted by publication date. Cat.: category; fb: feature-based; lb: label-based.}
\label{tab:transferability-metrics}
\footnotesize
\begin{tabular}{>{\raggedright}p{2cm}>{\raggedright}l>{\raggedright}p{3cm}>{\raggedright\arraybackslash}p{6.2cm}} \hline
Tr.~scorer & Cat. & Scorer input & Details \\ \hline
H-Score~\cite{bao2019information}   & fb & source feature extractor \& labels & transferability correlates to inter-class variance and feature redundancy \\ 
NCE~\cite{tran2019transferability}  & lb & source classification head \& labels & negative conditional entropy between source and target labels \\ 
LEEP~\cite{nguyen2020leep} & lb & source classification head \& labels & log-likelihood between target labels and source model predictions\\ 
N-LEEP~\cite{li2021ranking} & fb & source feature extractor \& labels  & log-likelihood between target labels and Gaussian mixture model fit to target extracted features \\ 
LogME~\cite{you2021logme} & fb & source feature extractor \& labels & probability of target labels conditioned on target image embeddings \\ 
Regularized H-Score~\cite{ibrahim2021newer} & fb & source feature extractor \& labels & shrinkage estimators for stable covariance  \\ 
GBC~\cite{pandy2022transferability} & fb &  source feature extractor \& labels &  Bhattacharyya coeff. between multivariate Gaussians fit to each class’ feature estimating overlap with target task classes \\ \hline
\end{tabular}
\end{table}

Ibrahim et al.\cite{ibrahim2021newer} and Agostinelli et al.\cite{agostinelli2022stable} evaluated transferability scores on general-purpose datasets for classification and segmentation tasks. Their findings suggest that these scores may be unstable, and minor variations in the experimental protocol could lead to different conclusions. N-LEEP and LogME deliver the best transferability estimation results depending on the experimental design of classification tasks. Our work focuses on classification tasks in scenarios where the dataset shift is significant. The experimental design of previous works assumes a lower dataset shift compared to what we consider in our paper. For instance, transferring from ImageNet to CIFAR is expected to be easier than any medical dataset due to some overlap between target-source classes and features. Additionally, we perform thorough hyperparameter tuning, which is essential in these works.

\section{Materials and Methods}
\label{sec:experiments}

\subsection{Datasets}

We assess three medical classification problems. We use ISIC2019~\cite{codella2018skin} for melano\-ma vs.~benign classification task and PAD-UFES-20~\cite{pacheco2020pad} for out-of-distribution (OOD) evaluation. BreakHis~\cite{breakhis} is used for histopathology breast cancer malign vs.~benign sample classification and ICIAR2018~\cite{rakhlin2018deep} for out-of-distribution assessment. For brain tumor classification, we use BrainTumor-Cheng~\cite{cheng2015enhanced}, a four-class dataset of MRI images. We adopt the NINS~\cite{brima2021deep} as the out-of-distribution test dataset.

\subsection{Methodology}

We aim to provide a fair and concise evaluation of each transferability scoring method described in Section~\ref{sec:concepts}. We restrict our analysis to pre-trained models on the ImageNet dataset. 
We focus exclusively on the \textit{source model selection} scenario, which aims to identify the most suitable pre-trained model for a given target dataset. Our methodology involves the following seven steps:
\begin{enumerate}
    \item Choosing a target medical task $T$.
    \item Selecting a pre-trained model architecture $A$.
    \item Computing the in-distribution transferability score $S_\text{id}(M,T,A)$ for all transferability scoring methods $M$, pre-trained model $A$, and the training dataset of task $T$.
    \item Performing a traditional evaluation of architecture $A$ for the target task $T$, by first optimizing the model hyperparameters on $T$'s validation dataset using the target metric to obtain the best model $A_\text{opt}(T)$, and then evaluating that metric $P_\text{id}(T,A)$ on $T$'s in-distribution test dataset.
    \item Computing the out-of-distribution transferability score $S_\text{ood}(M,T,A)$ for all transferability scoring methods $M$, the fine-tuned model $A_\text{opt}(T)$ obtained in the previous step, and target task $T$'s out-of-distribution test dataset (as explained in the previous subsection).
    \item Evaluating the target metric $P_\text{ood}(T,A)$ of $A_\text{opt}(T)$ on $T$'s out-of-distribution test dataset.
    \item For a given dataset $T$ and scoring method $M$, once steps 1-6 have been performed for all architectures, we may compute the correlation index between the transferability scores $S_*(M,T,A)$ and the traditional empirical performance metrics $P_*(T,A)$ across all architectures $A$. We showcase each one of those correlation analyses on a separate subplot of our results.  
\end{enumerate}

In our experiments, the target metric is always the balanced accuracy, and the correlation index is always the Kendall's tau, which ranges between $-$1 and 1, with positive correlations indicating higher-quality scoring methods. Zero correlations indicate that the scoring method has no ability to predict transferability. Negative correlations are harder to interpret: although they suggest predictive ability, they show the scoring method is working \textit{against} its expected design.

We analyze separately the in-distribution and the out-of-distribution analyses. As far as we know, we are the first to attempt OOD analyses on transferability metrics.

\section{Results}
\label{sec:results}

\noindent \textbf{\textit{Models architectures \& hyperparameter tuning}.}
\label{subsec:model-tuning}
We use 10 ImageNet pre-trained models:  
ResNet18~\cite{he2016deep}, ResNet34~\cite{he2016deep}, ResNet50 \cite{he2016deep}, MobileNetV2-0.5~\cite{sandler2018mobilenetv2}, MobileNetV2-1.0~\cite{sandler2018mobilenetv2}, DenseNet121~\cite{huang2017densely}, DenseNet161 \cite{huang2017densely}, DenseNet169~\cite{huang2017densely}, EfficientNet-B0~\cite{tan2019efficientnet}, ViT-Small~\cite{dosovitskiy2020image}.

For hyperparameter tuning, we followed the Tuning Playbook~\cite{tuningplaybookgithub} guidelines, using Halton sequences~\cite{halton_sequence} to sample candidates for the hyperparameters of interest. In our search, we keep fixed the use of SGD as the optimizer, cosine scheduler, 100 epochs, and batch size of 128. We search over $75$ quasi-random combinations of learning rate in range [$10^{-4}, 10^{-1}$] and weight decay in range [$10^{-6}, 10^{-4}$] for each model architecture, as those are the two most critical optimization hyperparameters~\cite{lirethinking}. 
We run the experiments on NVIDIA RTX 5000, RTX 8000.  
We select the best-performing model in the validation set for each architecture for test evaluation. In total, we trained $2250$ models. The source code to reproduce our experiments is available at \url{https://github.com/VirtualSpaceman/transfer-estimation-medical}.

\begin{figure}[t]
    \centering
    \includegraphics[width=\textwidth]{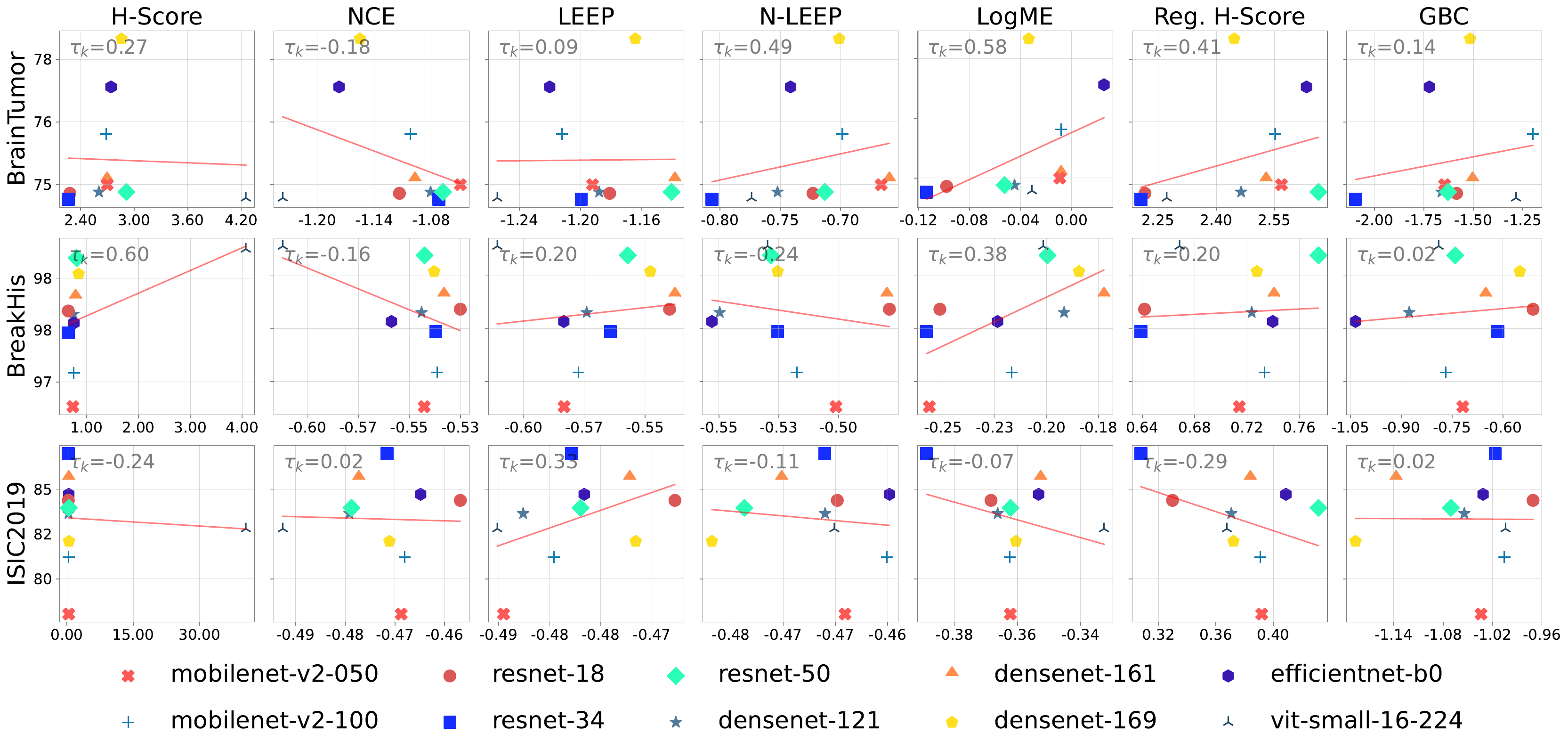}
    \caption{Evaluation of all scores (columns) and medical datasets (rows), showcasing the correlation between transferability scores (x-axis) and best accuracy on test (y-axis). The linear regression lines (in red) are for illustration only, as the correlation index employed is the non necessarily linear Kendall's tau, shown inside of each plot, on the top-left corner.}
    \label{fig:dataset-by-dataset-results}
\end{figure}

\vspace{0.15cm}
\noindent \textbf{\textit{In-distribution}.}
Fig.~\ref{fig:dataset-by-dataset-results} shows the results for each transferability scoring method and each model's architecture for all medical tasks. The red line indicates a regression line to show any tendency in the results. Table~\ref{tab:results-dataset-by-dataset} shows all investigated transferability scores for brain tumor, histopathologic, and skin lesion classification tasks, respectively. Each row depicts one correlation index value for that transferability scoring methods (columns). We calculate each correlation index considering the test performance of the best-fine-tuned model and the estimated transferability score for each architecture.

Methods such as LogME, N-LEEP, and NCE demonstrate varying degrees of correlation, indicating their potential as indicators of transferability within the same distribution. All transferability scoring methods exhibited an unstable behavior, as the correlation index consistently shifted across datasets. While LogME was one of the best methods for the BrainTumor-Cheng and Break\-His datasets, it exhibited negative correlations for ISIC2019. To our knowledge, such instability has not been observed in general-purpose computer vision datasets. Our main hypothesis for this phenomenon relates to the dissimilarity between source and target domains. Unlike general-purpose computer vision datasets, which often overlap in target and label sets or share similar features, medical transfer learning involves substantial domain differences.

\vspace{0.15cm}
\noindent \textbf{\textit{Out-of-distribution}.} It is easy to come across out-of-distribution sets in the real world, as deep learning datasets often exhibit diversity and correlation shifts~\cite{ye2022ood}. We conducted additional experiments to evaluate transferability scores' ability to predict out-of-distribution performance.

Table~\ref{tab:results-dataset-by-dataset} shows the transferability scores and correlation indexes, with interestingly high predictive capabilities observed for label-based transferability scoring methods. NCE and LEEP exhibited outstanding results for both ICIAR2018 and PAD-UFES-20 datasets across all correlations, with NCE being favored over other methods. We hypothesize that label-based methods are prone to provide better results than feature-based for binary tasks in out-of-distribution scenarios. As the number of classes of source dataset matches the target one, the probabilities distributions tend to concentrate on a single class, inflating the transferability score for binary cases.

\begin{table}[tb]
\centering
\caption{Kendall's tau ($\tau_{w}$) correlation index for each transferability scorer considering in and out-of-distribution scenarios for each medical task. }
\label{tab:results-dataset-by-dataset}
\resizebox{\columnwidth}{!}{%
\begin{tabular}{c|c|c|c|c|c|c|c|c}
\hline
~Task~ &
  ~Dataset~ &
  ~H-Score~ &
  ~NCE~ &
  ~LEEP~ &
  ~N-LEEP~ &
  ~LogME~ &
  ~Reg.~H-Score~ &
  ~GBC~ \\ 
  \hline
\multirow{2}{*}{Brain Tumor}                   
                                                        & BrainTumor-cheng & 0.270  & -0.180  & 0.090 & 0.494  & 0.584  & 0.405  & 0.135  \\ 
                                                        & NINS  & -0.333  & 0.156 & 0.200 & -0.333 & -0.289 & -0.422 & 0.200  \\ \hline
\multirow{2}{*}{Histopathologic}            
                                                        & BreakHis & 0.600  & -0.156 & 0.200 & -0.244 & 0.378  & 0.200  & 0.022   \\
                                                        & ICIAR2018  & 0.333   & 0.778 & 0.778 & 0.289  & 0.289  & 0.378 & 0.156  \\
                                                         \hline
\multirow{2}{*}{Skin Lesion}                                
                                                        & ISIC2019 & -0.244 & 0.022  & 0.333 & -0.111 & -0.067 & -0.289 & 0.022   \\
                                                        & PAD-UFES-20  & -0.156  & 0.911 & 0.422 & -0.156 & -0.022 & -0.022 & 0.067  \\
                                                        \hline
\end{tabular}
}
\end{table}

\vspace{0.15cm}
\noindent \textbf{\textit{Hypotheses why metrics failed}.} Up to this point, our experiments revealed that all transferability scoring methods present unstable quality. For example, both NCE and LEEP excel at out-of-distribution but report poor results in in-distribution scenarios. We hypothesize two factors that may contribute to the failure of the methods followed by some preliminary experiments: 
1) domain shift: the domain difference between source and target datasets might cause the failure. We fine-tuned each model on each medical dataset and evaluated their transferability score to the validation set. Our experiment indicates that only label-based methods excel in this scenario. So, domain shift helps to degrade the efficiency of such scores, but it is not the main reason. 
2) number of classes: to measure the impact of the number of classes in the transferability scores, we take the OxfordPets dataset and map the original 37 classes (dogs and cats breeds) into a binary problem (cats vs.~dogs). Our preliminary results suggest that all correlation indexes decrease, but all metrics still present high transferability estimation capabilities.  

\section{Conclusion}
\label{sec:conclusion}

Our work is the first to investigate the quality of transferability scoring methods for medical applications. We evaluated $7$ different transferability scoring methods in $3$ medical classification datasets, considering $10$ different architectures. Despite promising results in our out-of-distribution experiment, the instability presented by the scores across datasets in the in-distribution scenario lead us to recommend to practitioners not yet to rely on transferability scores for source model selection in medical image analysis. Our work takes one step towards reducing the need for expensive training by selecting pre-trained models efficiently that empowers the final performance on the target task. Such efficiency positively diminishes the carbon footprint when performing a hyperparameter search using a subset of deep learning architectures instead of all available.

Label-based methods shows superior results in out-of-distribution scenarios. Out-of-distribution scores might be inflated for binary tasks due to the distribution concentration on a single class, and the low number of classes benefits in favor of high transferability scores. Such an issue is absent in the available benchmarks because the general-purpose classification datasets present many classes and consider transferring from ImageNet as standard practice.

For future work, the analysis can be expanded to other configurations, such as finding the most related target task for a given source model (target model's selection) or cross-dataset transfer evaluation. Finally, evaluating future transferability scorers should include contexts where the difference between source and target domains is high, such as medical. This brings opportunities to assess the robustness of transferability scores regarding a limited amount of samples, unbalanced labels, and low inter- and high intra-variation classes.

\vspace{0.25cm}
\noindent\textbf{\textit{Data Use.}} We use only publicity available medical datasets, including PAD-UFES-20~\cite{pacheco2020pad}, ICIAR\-2018 \cite{rakhlin2018deep}, BreakHis~\cite{breakhis}, BrainTumor-Cheng~\cite{cheng2015enhanced}, NINS~\cite{brima2021deep}, and ISIC2019~\cite{codella2018skin}. All of them are under CC BY 4.0 license, except ISIC2019 (CC BY-NC 4.0). The data collection process is described in the original papers. 

\vspace{0.25cm}
\noindent\textbf{\textit{Acknowledgments}.} L. Chaves is funded by Becas Santander/Unicamp – HUB 2022, Google LARA 2021, in part by the Coordenação de Aperfeiçoamento de Pessoal de Nível Superior – Brasil (CAPES) – Finance Code 001, and FAEPEX. A.~Bissoto is funded by FAPESP (2019/19619-7, 2022/09606-8). S.~Avila is funded by CNPq 315231/2020-3, FAPESP 2013/08293-7, 2020/09838-0, H.IAAC, Google LARA 2021 and Google AIR~2022.

%
%
%
\bibliographystyle{splncs04}
\bibliography{ref}

\end{document}